\documentclass[10pt,twocolumn,letterpaper]{article}

\usepackage[pagenumbers]{cvpr} 

\usepackage{graphicx}
\usepackage{amsmath}
\usepackage{amssymb}
\usepackage{booktabs}
\usepackage{arydshln}
\usepackage{booktabs}
\usepackage{algorithm}
\usepackage{algorithmic}

\usepackage[pagebackref,breaklinks,colorlinks]{hyperref}

\usepackage[capitalize]{cleveref}
\crefname{section}{Sec.}{Secs.}
\Crefname{section}{Section}{Sections}
\Crefname{table}{Table}{Tables}
\crefname{table}{Tab.}{Tabs.}

\def\cvprPaperID{6344} 
\def\confName{CVPR}
\def\confYear{2022}

\begin{document}

\title{TeachAugment: Data Augmentation Optimization Using Teacher Knowledge}

\author{Teppei Suzuki \vspace{.5em}\\
Denso IT Laboratory, Inc.\\
}
\maketitle
\renewcommand{\thefootnote}{}
\begin{abstract}
Optimization of image transformation functions for the purpose of data augmentation has been intensively studied.
In particular, adversarial data augmentation strategies, which search augmentation maximizing task loss, show significant improvement in the model generalization for many tasks.
However, the existing methods require careful parameter tuning to avoid excessively strong deformations that take away image features critical for acquiring generalization.
In this paper, we propose a data augmentation optimization method based on the adversarial strategy called TeachAugment, which can produce informative transformed images to the model without requiring careful tuning by leveraging a teacher model.
Specifically, the augmentation is searched so that augmented images are adversarial for the target model and recognizable for the teacher model.
We also propose data augmentation using neural networks, which simplifies the search space design and allows for updating of the data augmentation using the gradient method.
We show that TeachAugment outperforms existing methods in experiments of image classification, semantic segmentation, and unsupervised representation learning tasks.\footnote[0]{Code: \url{https://github.com/DensoITLab/TeachAugment}}
\end{abstract}
\renewcommand{\thefootnote}{\arabic{footnote}}

\section{Introduction}
\label{sec:intro}
Data augmentation is an important technique used to improve model generalization.
To automatically search efficient augmentation strategies for model generalization, AutoAugment\cite{AA} has been proposed.
Searched data augmentation policies lead to significant generalization improvements.
However, AutoAugment requires thousands of GPU hours to search for efficient data augmentation.

Recent studies\cite{FastAA,FasterAA,pba} have demonstrated methods leading to dramatic reductions in search costs with AutoAugment, meaning that computational costs are no longer a problem.
In particular, online data augmentation optimization frameworks\cite{AdvAA,PA,onlineaugment,madao} that alternately update augmentation policies and a target network have not only reduced the computational costs but also simplified the data augmentation search pipeline by unifying the search and training processes.
\begin{figure}
    \centering
    \includegraphics[clip,width=0.9\hsize]{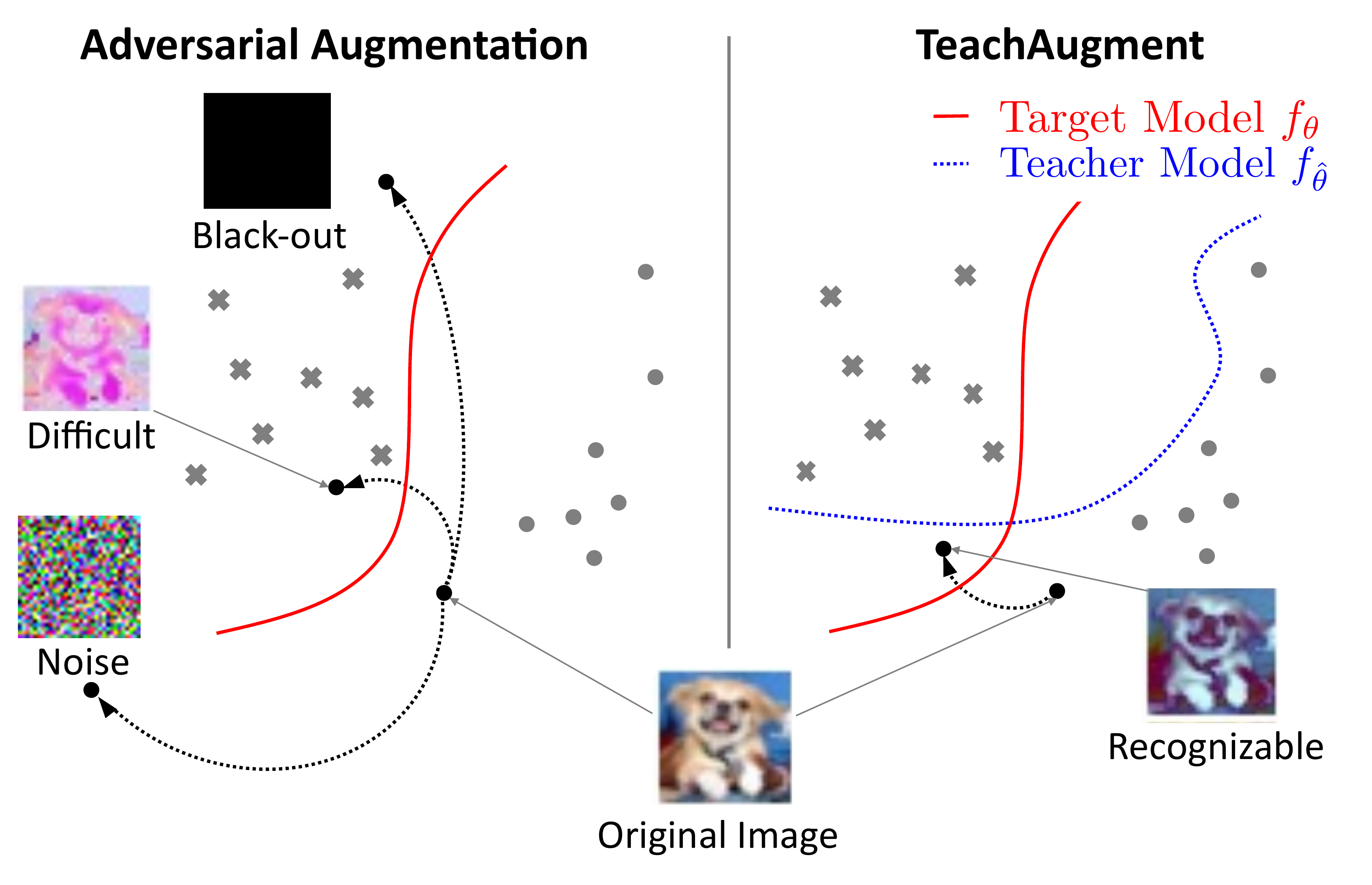}
    \caption{Concept of TeachAugment.
    Adversarial data augmentation, a baseline method, transforms data to increase loss for the target model $f_\theta$.
    The augmented data are often meaningless (e.g., black-out and noise images) or difficult to recognize without any constraints.
    TeachAugment, the proposed method, transforms data so that they are adversarial for the target model but they are recognizable for the teacher model $f_{\hat{\theta}}$. As a result, the augmented images will be more informative than the adversarial data augmentation.}
    \label{fig:fig1}
\end{figure}

Many online optimization methods are based on an adversarial strategy that searches augmentation maximizing task loss for the target model, which is empirically known to improve model generalization\cite{AdvAA,vat,rat,adv-pose,PA}.
However, the adversarial data augmentation is unstable without any constraint because maximizing the loss can be achieved by collapsing the inherent meanings of images, as shown in Fig. \ref{fig:fig1}.
To avoid the collapse, previous methods regularize augmentation based on prior knowledge and/or restrict the search range of the magnitude parameters of functions in the search space, but there are many tuned parameters that will annoy practitioners.

To alleviate the parameter tuning problem, we propose an online data augmentation optimization method using teacher knowledge, called TeachAugment.
TeachAugment is also based on the adversarial data augmentation strategy, but it searches augmentation in the range where the transformed image can be recognizable for a teacher model, as shown in Fig. \ref{fig:fig1}.
Unlike previous adversarial data augmentation methods~\cite{AdvAA,onlineaugment,vat,rat}, thanks to the teacher model, TeachAugment does not require priors and hyperparameters to avoid the excessively strong augmentation that collapses the inherent meanings of images.
As a result, TeachAugment does not require parameter tuning to ensure that the transformed images are recognizable.

Moreover, we propose data augmentation using neural networks that represent two functions, geometric augmentation and color augmentation.
Our augmentation model applies only two transformations to data but they can represent most functions included in the search space of AutoAugment and their composite functions.
The use of the neural networks has two advantages compared to conventional augmentation functions: (i) we can update the augmentation parameters using the gradient method with backpropagation, and (ii) we can reduce the number of functions in the search space from tens of functions to two functions.
In particular, because of the latter advantage, practitioners only need to consider the range of the magnitude parameters for the two functions when they adjust the size of the search space for better convergence.

\textbf{Contribution.}
Our contribution is summarized as follows:
(1) We propose an online data augmentation optimization framework based on the adversarial strategy using teacher knowledge called TeachAugment.
TeachAugment makes the adversarial augmentation more informative without careful parameter tuning by leveraging the teacher model to avoid collapsing the inherent meaning of images.
(2) We also propose data augmentation using neural networks. The proposed augmentation simplifies the search space design and enables updating of its parameters by the gradient method in TeachAugment.
(3) We show that TeachAugment outperforms previous methods, including online data augmentation\cite{PA,AdvAA} and state-of-the-art augmentation strategies\cite{randaug,trivialaug} in classification, semantic segmentation, and unsupervised representation learning tasks without adjusting the hyperparameters and size of the search space for each task.

\section{Related work}
As conventional data augmentation for image data, geometric and color transformations are widely used in deep learning.
Besides, advanced data augmentations\cite{bclearn,randomerasing,augmix,samplepairing,mixup,cutout,cutmix} have been recently developed and they have improved accuracy on image recognition tasks.
Data augmentation not only enhances the image recognition accuracy, but also plays an important role in recent unsupervised representation learning\cite{moco,mocov2,byol,simclr,simsiam} and semi-supervised learning\cite{vat,rat,fixmatch}.
While the data augmentation usually improves model generalization, it sometimes hurts performance or induces unexpected biases.
Thus, one needs to manually find the effective augmentation policies based on domain knowledge to enhance model generalization.

Cubuk et al.\cite{AA} proposed a method to automatically search for effective data augmentation, called AutoAugment.
AutoAugment outperforms hand-designed augmentation strategies and shows state-of-the-art performances on various benchmarks.
Data augmentation search has become a research trend, and many methods have been proposed\cite{FastAA,FasterAA,dada,AdvAA,aws,selfaugment,PA,onlineaugment,ohl,trivialaug,randaug,uniformaugment,madao,wei2020circumventing}.

We roughly categorize them into two types: a \textit{proxy task based} method and a \textit{proxy task free} method.
The proxy task based methods\cite{FastAA,FasterAA,dada,AdvAA,aws} search data augmentation strategies on \textit{proxy tasks} that use subsets of the training data and/or small models to reduce computational costs.
Thus, policy searches using the proxy task based method might be sub-optimal.
The proxy task free methods~\cite{PA,uniformaugment,trivialaug,AdvAA,randaug} directly search data augmentation strategies on the target network with all training data.
Thus, policies obtained with this method are potentially optimal.

In proxy task free methods, several approaches, such as RandAugment~\cite{randaug} and TrivialAugment~\cite{trivialaug}, randomize the parameters searched and reduce the size of the search space.
Other methods, such as Adversarial AutoAugment~\cite{AdvAA} and PointAugment~\cite{PA}, update augmentation policies in an online manner, meaning that they alternately update a target network and augmentation policies.
The online optimization methods simplify data augmentation optimization frameworks by unifying the search and training processes.
However, these methods are fraught with minor problems.
For example, PointAugment~\cite{PA} unjustly binds the difficulty of augmented images, Adversarial AutoAugment~\cite{AdvAA} manually restricts the search space to guarantee convergence, and OnlineAugment~\cite{onlineaugment} has many hyperparamters for regularization.
These problems stem from reliance on the adversarial strategy that searches augmentation maximizing task loss.
In other words, these problems are induced to ensure that the transformed images are recognizable.

In this work, we focus on proxy task free methods updating the policies in an online manner for two reasons: (i) they can directly search data augmentation strategies on the target network with all training data, and (ii) they unify the search and training processes, simplifying the framework.

\section{Data augmentation optimization using teacher knowledge}
\begin{figure*}
    \centering
    \includegraphics[clip,width=0.8\hsize]{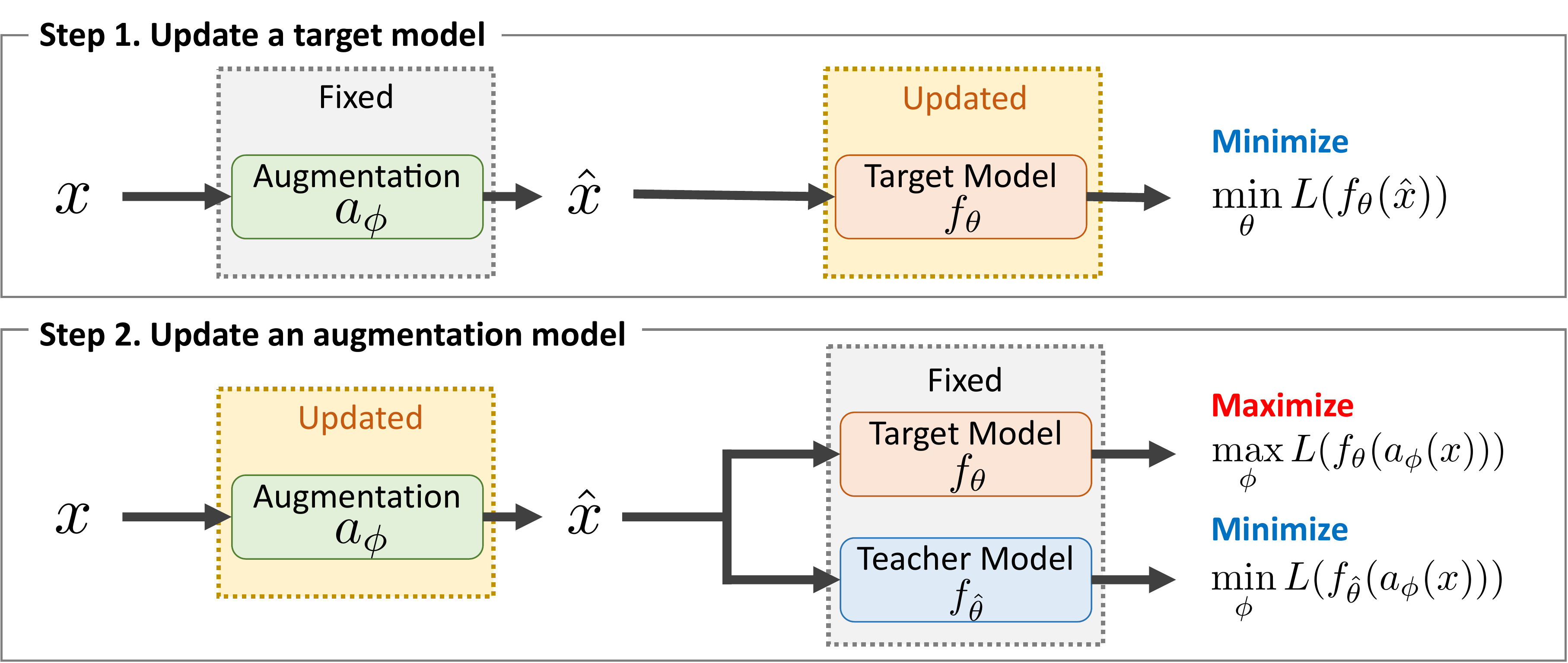}
\caption{Overview of training procedure for TeachAugment. Our method alternately updates the target model $f_\theta$ and the augmentation model $a_\phi$ (i.e., step 1 and step 2 are repeated).}
    \label{fig:overview}
\end{figure*}
\subsection{Preliminaries}
Let $x\sim\mathcal{X}$ and $a_\phi$ be an image sampled from dataset $\mathcal{X}$ and an augmentation function parameterized by $\phi$, respectively.
In conventional data augmentation, $\phi$ corresponds to the magnitude of augmentation, while in this work, it corresponds to the neural network’s parameters.
In general training procedures using data augmentation, the mini-batch samples are transformed by $a_\phi$ and fed into a target network $f_\theta$ (i.e., $f_\theta(a_\phi(x))$).
Then, the parameters of the target network are updated to minimize task loss $L$ in the stochastic gradient descent.
This training procedure is represented as $\min_\theta \mathbb{E}_{x\sim\mathcal{X}}L(f_\theta(a_\phi(x))$.
Note that we omitted the target label because we consider not only supervised learning but also unsupervised representation learning in this work.

In addition, adversarial data augmentation searches the parameters $\phi$ maximizing the loss.
The objective is defined as $\max_\phi\min_\theta \mathbb{E}_{x\sim\mathcal{X}}L(f_\theta(a_\phi(x))$.
This objective is sometimes solved by alternately updating $\phi$ and $\theta$\cite{AdvAA,PA}.
The adversarial data augmentation is empirically known to improve model generalization\cite{AdvAA,adv-pose}.
However, it does not work without some regularization or restriction in the size of the search space because the maximization with respect to $\phi$ can be achieved by collapsing the inherent meanings of $x$.
Thus, instead of using regularization based on prior knowledge, we utilized a teacher model to avoid the collapse.

\subsection{TeachAugment}
Let $f_{\hat{\theta}}$ be the teacher model, which allows for any model as long as it is different from the target model $f_\theta$.\footnote{In most of our experiments, we define the teacher model as the same model as the target model but with different parameters. Thus, the same symbol $f$ is used for the teacher model and the target model although it may be a little complicated.}
In this work, we suggest two types of the teacher model, a pretrained teacher and an EMA teacher whose weights are updated as an exponential moving average of the target model's weights.
We provide detailed definitions and evaluate the effect of teacher model choices in Sec. \ref{sec:ablation}.
 
The proposed objective is defined as follows:
\begin{align}
    \label{eq:obj}
    \max_\phi\min_\theta \mathbb{E}_{x\sim\mathcal{X}}\left[L(f_\theta(a_\phi(x))) - L(f_{\hat{\theta}}(a_\phi(x)))\right].
\end{align}
For the target model, this objective has the same properties as the adversarial data augmentation, but the augmentation function needs to minimize the loss for the teacher model, in addition to maximizing the loss for the target model.
The augmentation that is obtained in this manner avoids collapsing the inherent meanings of images because the loss for the teacher model will explode when the transformed images are unrecognizable.
In other words, the introduced teacher loss requires the augmentation function to transform images so that they are adversarial for the target model in the range where they are recognizable for the teacher model.

As shown in Fig. \ref{fig:overview}, the objective is solved by alternately updating the augmentation function and the target model in the stochastic gradient descent, similar to previous methods~\cite{madao,PA,AdvAA}.
We first update the target network for $n_\text{inner}$ steps, and then update the augmentation function.
A pseudo-code can be found in the appendix.
Note that the augmentation function is updated by the gradient method because our proposed augmentation using the neural networks introduced in Sec. \ref{sec:augmentation} is differentiable with respect to $\phi$.
We refer to the augmentation strategy following Eq. \ref{eq:obj} as \textit{TeachAugment}.

Unlike previous methods\cite{PA,onlineaugment}, TeachAugment does not regularize augmentation functions based on the domain knowledge, such as cycle consistency and smoothness\cite{onlineaugment}. It also does not bind the difficulty of the transformed images as in \cite{PA} to ensure that the transformed images are recognizable.

\subsection{Improvement techniques}
\label{sec:tech}
The training procedure for TeachAugment is similar to that for generative adversarial networks (GANs)\cite{gan} and actor-critic methods in reinforcement learning\cite{Sutton2000policy,konda2000actor}, which alternately update two networks.
Practitioners in both fields have amassed a large number of strategies to mitigate instabilities and improve training\cite{pfau2016connecting}.
TeachAugment also benefits from three techniques used in both fields, \textit{experience replay}, \textit{non-saturating loss}, and \textit{label smoothing}.
Moreover, we introduced color regularization to mitigate inconsistency between color distributions of images before and after data augmentation.
The techniques introduced here are also applicable for other online methods\cite{PA,AdvAA,onlineaugment}.

\textbf{Non-saturating loss.}
For classification tasks, the loss function $L$ is usually defined as the cross-entropy loss, $L(f_\theta(a_\phi(x))=\sum_{k=1}^K-y_k\log f_\theta(a_\phi(x))_k$, where $y\in \{0,1\}^K$ and $K$ denote the one-hot ground truth label and the number of classes, respectively.
In this case, the gradient of the first term in Eq. \eqref{eq:obj} often saturates in the maximization problem with respect to $\phi$ when the target model's predictions are very confident.
Thus, we use $\sum_{k=1}^Ky_k\log (1-f_\theta(a_\phi(x))_k)$ when updating the augmentation model rather than $\sum_{k=1}^K-y_k\log f_\theta(a_\phi(x))_k$.
This technique has been used in GANs\cite{gan}.

The non-saturating loss is a key factor for TeachAugment; it improves error rates of WideResNet-28-10\cite{wrn} on CIFAR-100\cite{cifar} from 18.7\% to 17.4\% (a baseline's error rate is 18.4\%).
Thus, we basically use the non-saturating loss for updating augmentation models in our experiments.

\textbf{Experience replay.}
In reinforcement learning, experience replay\cite{er,rb} stores actions chosen by the actor in the past and reuses them to update the critic.
We apply this technique to our method by storing the augmentation models and prioritizing them in a manner similar to prioritized experience replay\cite{per}.
Then, the target network is updated using the augmentation model randomly sampled from the buffer following their priorities.

Let $p_i$ be a priority of the $i$-th stored augmentation model.
We compute the priority as $p_i = \gamma^{S-i}$, where $S$ denotes the number of augmentation models stored in the buffer, $\gamma$ denotes a decay rate, and we set $\gamma$ to 0.99.
In our experiments, we stored the augmentation model every $n_\text{buffer}$ epochs.
For image classification on CIFAR-10, CIFAR-100, and semantic segmentation, $n_\text{buffer}$ was set to 10, and for other tasks and datasets, it was set to 1.

\textbf{Label smoothing.}
Label smoothing is a technique that replaces the one-hot labels with $\hat{y}_k=(1-\epsilon)y_k+\epsilon/K$, where $\epsilon\in [0,1)$ is a smoothing parameter.
For our method, the label smoothing prevents gradients from exploding when the target model's predictions are very confident under the non-saturating loss.
In particular, such a situation tends to occur for easy tasks or strong target models.
To mitigate exploding gradients, we used the smoothed labels for the first term in Eq. \eqref{eq:obj} when updating the augmentation model.
Note that for a fair comparison, we did not apply it when updating the target model.

\textbf{Color regularization.}
In practice, the color augmentation model tends to transform the pixel colors outside the color distribution of the training data.
As a result, the augmented images will be out-of-distribution data, which may hurt the recognition accuracy for the in-distribution samples.
To align the color distributions before and after augmentation, we regularized the color augmentation model by introducing sliced Wasserstein distance (SWD)~\cite{swd} between pixel colors before and after the augmentation.
SWD is a variant of Wasserstein distance that represents a distance between two distributions.

We define the color regularization term as follows:
\begin{align}
    L_\text{color}(\{x^b\}_b^B,\{\tilde{x}^b\}_b^B)=\sum_i\mathrm{SWD}(\{x_i^b\}_b^B,\{\tilde{x}_i^b\}_b^B),
\end{align}
where $\{x_i^b\}_b^B$ denotes a set of $i$-th pixels of images in a mini-batch with a batch size of $B$, and $\{\tilde{x}^b\}_b^B$ denotes color-augmented images defined in Eq. \eqref{eq:color-aug}.
Because costs of computing SWD at each pixel position depends linearly on the image resolution, we can compute SWD for high-resolution images handled in semantic segmentation with low computational resources.
Then, in the stochastic gradient descent, the gradient with respect to $\phi$ in each iteration is represented as follows:
\begin{align}
    \nonumber
    \frac{\partial}{\partial\phi}\frac{1}{B}\sum_b^B\left[L(f_\theta(a_\phi(x^b))) - L(f_{\tilde{\theta}}(a_\phi(x^b)))\right]\\
    -\lambda L_\text{color}(\{x^b\}_b^B,\{\tilde{x}^b\}_b^B),
\end{align}
where $\lambda$ is a hyperparameter controlling the effect of the regularization that was set to 10 in our experiments.

\section{Data augmentation using neural networks}
\label{sec:augmentation}
We propose data augmentation using neural networks parameterized by $\phi$, which consists of two models, a color augmentation model $c_{\phi_c}$ and a geometric augmentation model $g_{\phi_g}$.
Thus, $a_\phi$ is defined as the composite function of $c_{\phi_c}$ and $g_{\phi_g}$, $a_\phi=g_{\phi_g}\circ c_{\phi_c}$, and the parameter $\phi$ is a set of $\phi_c$ and $\phi_g$, $\phi=\{\phi_c,\phi_g\}$.
\begin{figure}
    \centering
    \includegraphics[width=0.9\hsize]{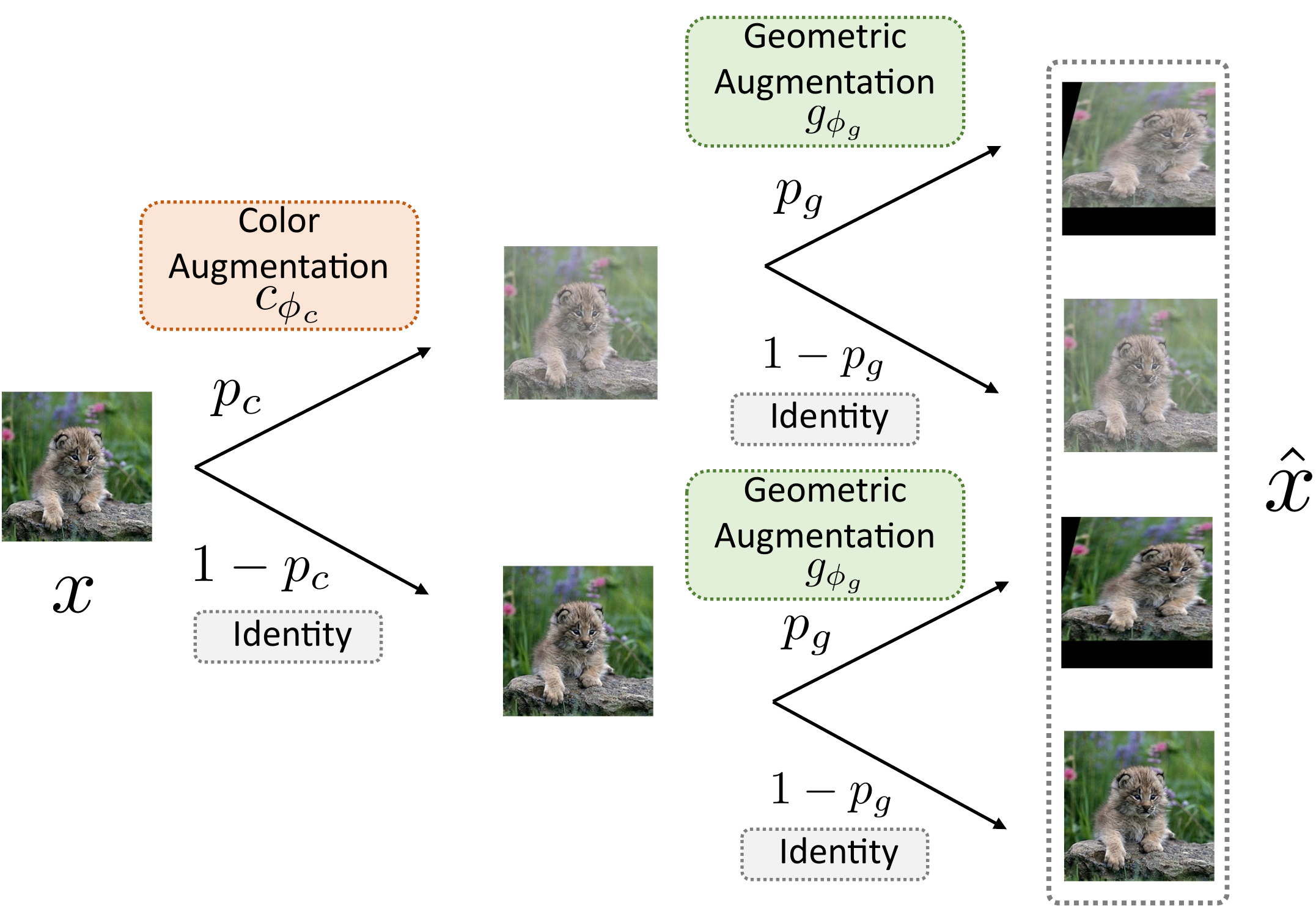}
    \caption{Data augmentation procedure. Our data augmentation consists of the color augmentation $c_{\phi_c}$ and the geometric augmentation $g_{\phi_g}$. Each augmentation is applied to an input image $x$ with probabilities $p_c$ and $p_g$.
    }
    \label{fig:aug}
\end{figure}
Our augmentation model enables updating of its parameters by the gradient method and construction of the search space with only two functions.

The augmentation procedure is illustrated in Fig. \ref{fig:aug}.
Given an image $x\in\mathbb{R}^{M\times 3}$, where $M$ denotes the number of pixels and 3 corresponds to the RGB color channels, the color augmentation is applied with probability of $p_c\in(0,1)$, and then the geometric augmentation is applied with probability of $p_g\in(0,1)$.

The color augmentation is defined as follows:
\begin{align}
    \label{eq:color-aug}
    \tilde{x}_i=t(\alpha_i\odot x_i+\beta_i),\ (\alpha_i, \beta_i) = c_{\phi_c}(x_i,z,c),
\end{align}
where $\alpha_i,\ \beta_i\in\mathbb{R}^3$ denote scale and shift parameters; $\odot$ denotes the element-wise multiplication between vectors; $t(\cdot)$ denotes the triangle wave, $t(x)=\arccos(\cos(x\pi))/\pi$, which ensures $\tilde{x}_i\in[0,1]$; $z\sim\mathcal{N}(0, I_N)$, where $\mathcal{N}(0, I_N)$ denotes $N$-dimensional unit Gaussian distribution, and $c$ is an optional context vector.
In our experiments, we used a one-hot ground truth label as $c$ for image classification and omitted it for other tasks.
The color augmentation model can transform the input image into any image \textit{in principle} when the augmentation model is sufficiently large because of the universal approximation theorem.

The geometric augmentation is defined as follows:
\begin{align}
    \label{eq:geo-aug}
    \hat{x} = \mathrm{Affine}(\tilde{x}, A+I),\ A = g_{\phi_g}(z,c),
\end{align}
where $\mathrm{Affine}(\tilde{x},A+I)$ denotes an affine transformation of $\tilde{x}$ with a parameter $A+I$; $I\in\mathbb{R}^{2\times 3}$ denotes a matrix where $\forall i, I_{ii}=1$ and 0 otherwise, which makes the affine transformation the identity mapping $\mathrm{Affine}(x,I)=x$; $A\in\mathbb{R}^{2\times 3}$ denotes the residual parameter, and $c$ and $z$ are the same vectors used in Eq. \eqref{eq:color-aug}.

The geometric augmentation model can be defined by transformations other than the affine transformation, similar to \cite{stn}.
However, the affine transformation can represent all geometric transformations in the search space of AutoAugment and their composite functions.
Thus, we considered only the affine transformation in this work.

In addition to $\phi_c$ and $\phi_g$, we also learned the probabilities $p_c$ and $p_g$ by using the gradient method.
However, the decision process of applying the augmentation is non-differentiable with respect to $p_c$ and $p_g$.
To make it differentiable with respect to the probabilities, we used the same approach as in previous works\cite{FasterAA,dada}.
A detailed pipeline can be found in the appendix.

\section{Experiments}
We evaluate our method with three tasks.
For an ablation study and comparison with existing automatic data augmentation search methods, we evaluate our method on image classification tasks.
We train WideResNet-40-2 (WRN-40-2)\cite{wrn}, WideResNet-28-10 (WRN-28-10)\cite{wrn}, Shake-Shake (26 2$\times$96d)\cite{shake}, and PyramidNet\cite{pyramidnet} with ShakeDrop regularization\cite{shakedrop} on CIFAR-10 and CIFAR-100\cite{cifar}, and train ResNet-50~\cite{resnet} on ImageNet\cite{imagenet}.
The training and evaluation protocols are the same settings as in previous work\cite{FastAA}.

In addition to the above experiments, we examine our method with semantic segmentation and unsupervised representation learning.
For semantic segmentation, we train FCN-32s~\cite{fcn}, PSPNet~\cite{pspnet}, and Deeplabv3~\cite{deeplabv3} on Cityscapes~\cite{cityscapes}.
The training protocol is the same as \cite{pspnet}.
For unsupervised representation learning, we pretrain ResNet-50~\cite{resnet} on ImageNet~\cite{imagenet} using SimSiam~\cite{simsiam} with various data augmentation and then evaluate the model following the linear evaluation protocol~\cite{simsiam}.

More details can be found in the appendix.

\subsection{Implementation details}
We construct the geometric augmentation model using a multi-layer perceptron (MLP) and the color augmentation model using two MLPs that receive an RGB vector and a noise vector as inputs and then add up the outputs.
We apply the sigmoid function to the outputs of each augmentation model and normalize the outputs in a range of $A\in(-0.25,0.25)^{2\times 3}$, $\alpha\in(0.6,1.4)$, and $\beta\in(-0.5,0.5)$.
The dimension of the noise vector $z$ is set to 128.
For stochasticity, Dropout~\cite{dropout} is applied after the linear layers, except to the output layer.
We initialize the weights of the output layer as zero to make the augmentation the identity mapping in the initial state.
We use AdamW optimizer~\cite{adamw} to train the augmentation model.
All the hyperparameters of AdamW (e.g., learning rate and weight decay) are set to the PyTorch default parameters~\cite{pytorch}.
More details can be found in the appendix.

\subsection{Ablation study}
\label{sec:ablation}
\textbf{Effect of teacher model choices.}
We first investigate the effect of teacher model choices.
We trained WRN-40-2 with two types of teacher models: a teacher that was pretrained with the baseline augmentation, and an EMA teacher that is exponential moving average of the target model (i.e., $\hat{\theta}\leftarrow\xi\hat{\theta}+(1-\xi)\theta$).
We set the decay rate $\xi$ to 0.999, following \cite{mean-teacher}.
For the pretrained teacher, we pretrained WRN-40-2, which is the same model as the target model, and WRN-28-10, which is a stronger model than the target model.

The results are shown in Tab. \ref{tab:ab_teacher}.
\begin{table}[t]
\scalebox{0.95}{
    \begin{tabular}{c|ccc}
        Teacher & WRN-40-2 & WRN-28-10 & EMA \\ \toprule[1.1pt]
        CIFAR-10      & 4.4          & \textbf{3.7}   & \textbf{3.7}\\
        CIFAR-100     & 21.3         & 21.6           & \textbf{20.3}
    \end{tabular}
    }
    \centering
    \caption{Effect of teacher models. We report the error rates (\%) of WideResNet-40-2 trained with various teacher models. WRN-40-2 and WRN-28-10 are pretrained with the baseline augmentation. EMA is an exponential moving average of the target model.}
    \label{tab:ab_teacher}
\end{table}
For both datasets, the EMA teacher achieves lower error rates than the others.
The augmentation model may cause overfitting to the teacher model when the pretrained teacher is used because they are not updated during training.
The EMA teacher would prevent overfitting by updating parameters at the same time as the target model during training, bringing more improvement than the pretrained teachers.

Interestingly, we found the stronger teacher was not a better teacher.
In fact, WRN-28-10 demonstrates error rates comparable to the EMA teacher in CIFAR-10, but in CIFAR-100, it was slightly worse than WRN-40-2.

In the rest of the experiments, we used the EMA teacher for TeachAugment because the EMA teacher not only has lower error rates but also eliminates the architecture choices of the teacher model.

\textbf{Effect of stabilizing techniques.}
We next investigate the effect of stabilizing techniques introduced in Sec. \ref{sec:tech}.
We trained WRN-28-10 on CIFAR-10 and CIFAR-100, with the exception of each technique.

The results are shown in Tab. \ref{tab:ab_technique}.
\begin{table}[]
    \centering
    \scalebox{0.95}{
    \begin{tabular}{l|cc}
        Dataset                       & CIFAR-10      & CIFAR-100 \\ \toprule[1.1pt]
        All                           & \textbf{2.5} & 16.8\\ \hline
        \quad w/o Color Reg.          & 3.0          & \textbf{16.5} \\ 
        \quad w/o Experience Replay   & 2.7          & 17.3 \\
        \quad w/o Label Smoothing     & 3.0          & 17.3 \\ \hline
        \quad w/o all techniques & 3.0          &    17.4
    \end{tabular}
    }
    \caption{Error rates (\%) of WideResNet-28-10 on CIFAR-10 and CIFAR-100 with various technique choices.}
    \label{tab:ab_technique}
\end{table}
Except for the color regularization on CIFAR-100, all of the techniques contributes to the improvement of error rates.

\begin{figure}
    \centering
    \begin{tabular}{ccc}
        \includegraphics[clip,width=0.25\hsize]{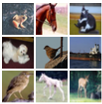} &
        \includegraphics[clip,width=0.25\hsize]{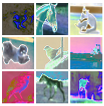} &
        \includegraphics[clip,width=0.25\hsize]{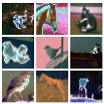}\\
    \end{tabular}
    \includegraphics[clip,width=0.9\hsize]{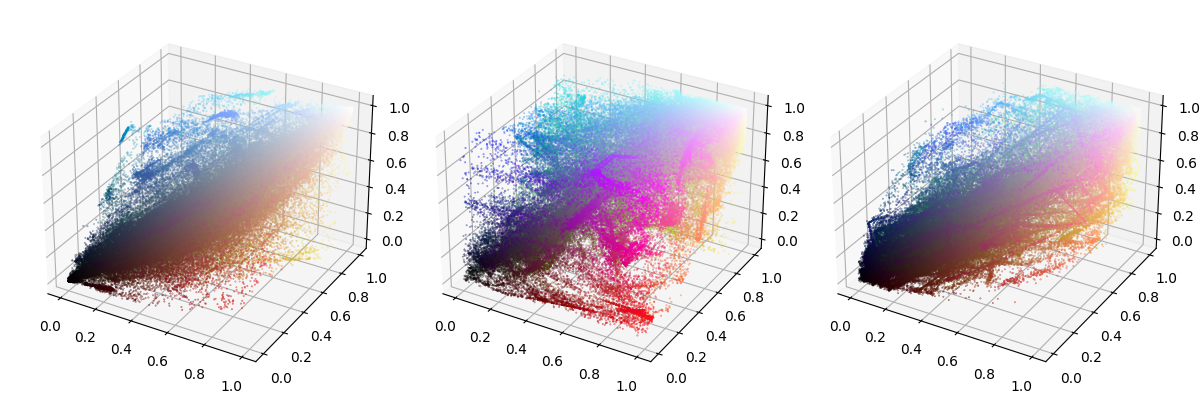}\\
    (a) CIFAR-10\\
    \vspace{3mm}
    \begin{tabular}{ccc}
        \includegraphics[clip,width=0.25\hsize]{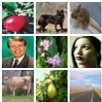} & 
        \includegraphics[clip,width=0.25\hsize]{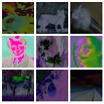} & 
        \includegraphics[clip,width=0.25\hsize]{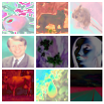}\\
    \end{tabular}
    \includegraphics[clip,width=0.9\hsize]{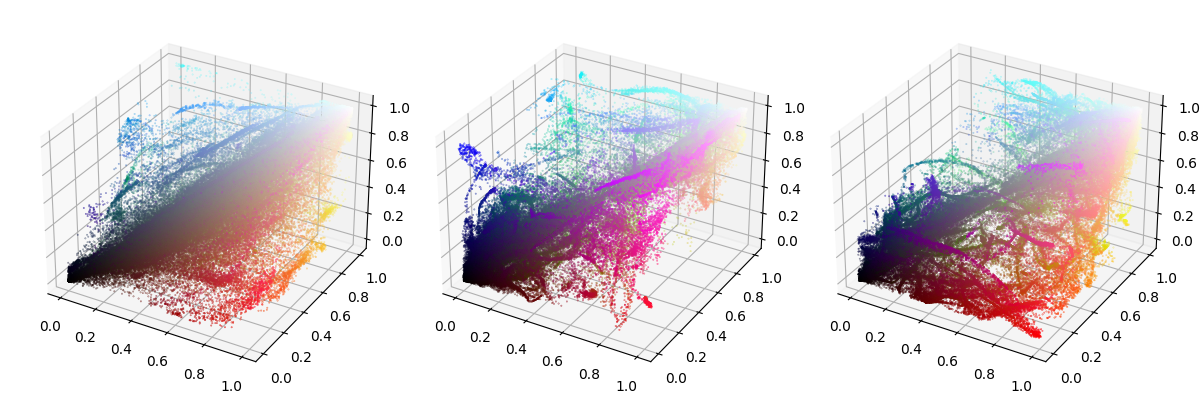}\\
    (b) CIFAR-100
    \caption{Augmented images and corresponding color distributions on CIFAR-10 and CIFAR-100. From left to right, original images, augmented images from the model trained without color regularization, and augmented images from the model trained with color regularization. We randomly picked 128 images and plotted their pixel colors (i.e., each point corresponds to a pixel).}
    \label{fig:ab_rgb}
\end{figure}

We visualize the effect of the color regularization in Fig. \ref{fig:ab_rgb}.
The color distributions are aligned by the color regularization.
In particular, the augmented images without regularization lose brightness in many pixels with CIFAR-100.
However, alignment of the color distribution improve the error rates only for CIFAR-10.
This would be due to the color diversity of CIFAR-10 being narrower than CIFAR-100, which can be seen from the color distributions in Fig. \ref{fig:ab_rgb}.
In other words, in the case of training without color regularization, the color augmentation produces out-of-distribution colors, and the obtained images hurt the recognition accuracy of the target model for the in-distribution samples.
The transformed image from CIFAR-10 tends to be this kind of an out-of-distribution sample due to the low diversity of colors.

\begin{figure}
    \centering
    \includegraphics[width=0.9\hsize,clip]{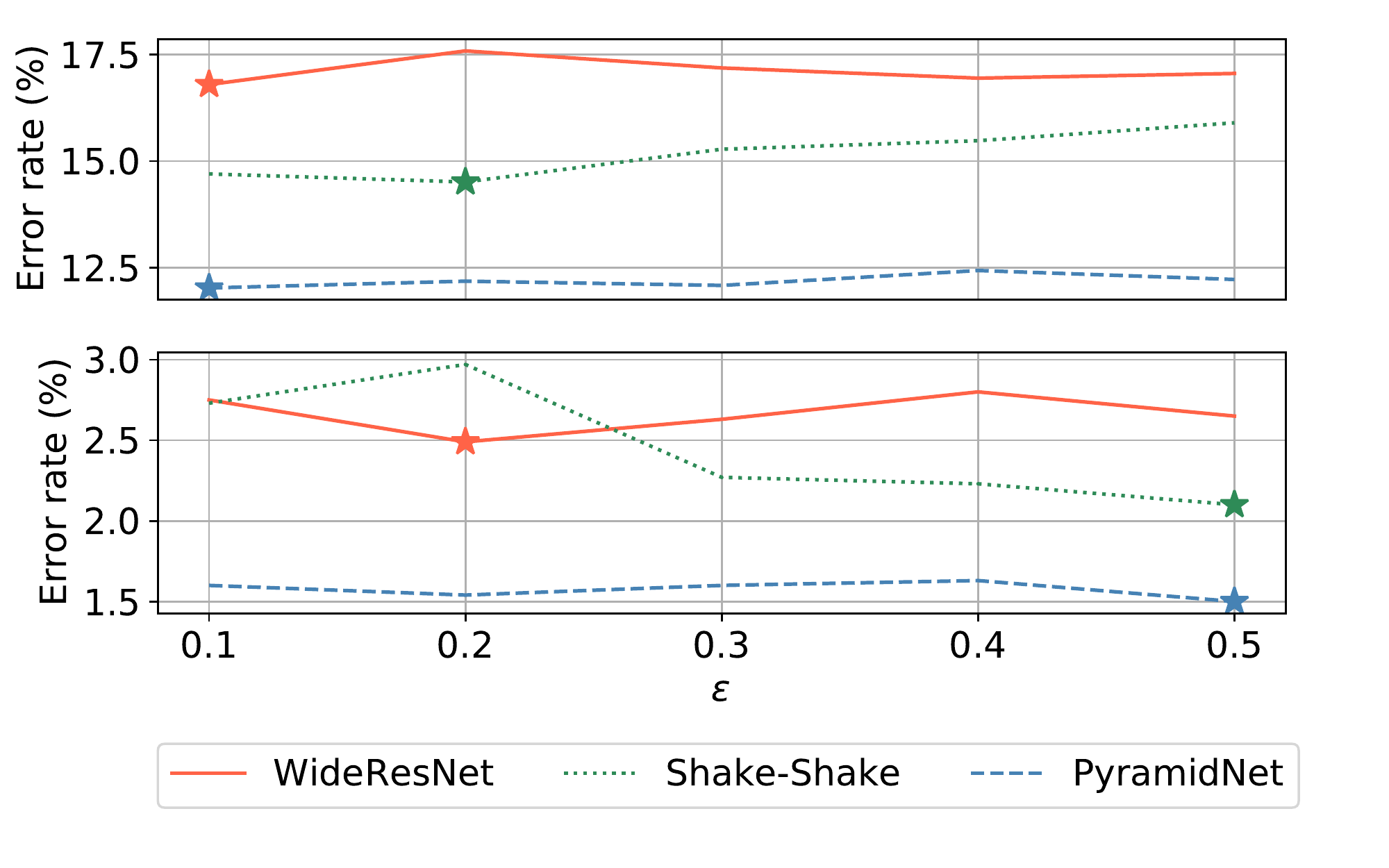}
    \caption{Effect of $\epsilon$ on label smoothing. The upper figure shows error rates on CIFAR-100, and the lower figure shows the error rates on CIFAR-10. Stars indicate $\epsilon$ achieving minimum error rates.}
    \label{fig:ab_lbl_smoothing}
\end{figure}

We evaluate label smoothing with various smoothing parameter values, and the results are shown in Fig. \ref{fig:ab_lbl_smoothing}.
For the easy tasks of CIFAR-10, label smoothing with the large $\epsilon$ works well.
As already described in Sec. \ref{sec:tech}, the gradient of the non-saturating loss tends to be large for easy tasks or strong models, meaning that the model's predictions tend to be very confident.
Thus, a larger $\epsilon$ avoids exploding gradients and stabilizes training for CIFAR-10.

\textbf{Evaluation of the proposed objective function.}
We evaluate the effectiveness of the proposed objective function, eq. \eqref{eq:obj}.
As baseline methods, we replace the objective function in our framework with the loss of adversarial AutoAugment (Adv. AA)~\cite{AdvAA} and PointAugment (PA)~\cite{PA}.
For a fair comparison, all of the models were trained with the same protocol and the difference between them was the objective function.
The detailed setting is described in the appendix.

The results are shown in Tab. \ref{tab:ab_obj}.
\begin{table}[t]
    \centering
    \scalebox{0.95}{
    \begin{tabular}{c|c|ccc}
    Objective & Baseline & Adv. AA~\cite{AdvAA} & PA~\cite{PA} & Ours \\ \toprule[1.1pt]
    CIFAR-10  &    3.1   & 3.6               &  2.9         & \textbf{2.5} \\
    CIFAR-100 &    18.4  & 19.4              &  17.5        & \textbf{16.8}
    \end{tabular}
    }
    \caption{Error rates (\%) of WideResNet-28-10 trained with various online optimization frameworks. All results are obtained under the same augmentation models and training protocol, except for the objective function.}
    \label{tab:ab_obj}
\end{table}
The error rates of Adv. AA degrades from the baseline because the augmentation model produces the unrecognizable images that confuse the target model.
For Adv. AA, one needs to carefully tune the size of the search space to guarantee the convergence\cite{AdvAA} but we do not adjust it.
In addition, for such instability, the proposed augmentation model would be unsuitable because of its property, i.e., the color augmentation model can transform the input image into any image.
In other words, the proposed augmentation would work only for the methods that guarantee transformed images to be recognizable.
Our method achieves better error rates than PointAugment.
PointAugment controls the upper bound of the loss for the augmented data with a dynamic parameter, but our method has no such restriction.
As a result, our method provides more diverse transformations and better improvements in the error rates than PointAugment.
The augmented images are visualized in the appendix.

\textbf{Evaluation of the proposed augmentation model.}
We compare the proposed augmentation model to a differentiable data augmentation pipeline (DDA) proposed in \cite{FasterAA} and augmentation models used in OnlineAugment (OA)\cite{onlineaugment}.
The results are shown in Tab. \ref{tab:dda}.
\begin{table}[]
    \centering
    \begin{tabular}{c|c|ccc}
        Augmentation & Baseline & DDA & OA & Ours \\ \toprule[1.1pt]
        CIFAR-10 & 3.1 & 2.8 & 16.7 & \textbf{2.5} \\
        CIFAR-100 & 18.4 & 18.0 & 26.2 & \textbf{16.8}
    \end{tabular}
    \caption{Error rates (\%) of WideResNet-28-10 with DDA~\cite{FasterAA}, the augmentation models of OnlineAugment~\cite{onlineaugment} (OA) and the proposed augmentation on CIFAR-10 and CIFAR-100. Each augmentation is optimized by the TeachAugment strategy.}
    \label{tab:dda}
\end{table}
The improvement of DDA from baseline is less than that of ours.
DDA uses some techniques (e.g., relaxation\cite{concrete,gsoftmax} and straight-through estimator\cite{ste}) to make some data augmentation functions differentiable, but these techniques induce inaccurate gradients and the vanishing gradient problem that would make it difficult to learn effective augmentation in the TeachAugment framework.
OA is much worse than baseline.
OnlineAugment regularizes augmentation models based on prior knowledge, such as cycle consistency and smoothness, instead of bounding the search space.
However, TeachAugment does not impose strong regularization.
The lack of regularization would be unsuitable for augmentation with the unbounded search space.
We note that the proposed augmentation is better than DDA and OA for TeachAugment, but as seen in Tab. \ref{tab:ab_obj}, it does not work in all cases.

\subsection{Image classification}
We compare our method to several previous methods: AutoAugment (AA)~\cite{AA}, PBA\cite{pba}, Fast AA\cite{FastAA}, Faster AA\cite{FasterAA}, DADA\cite{dada}, RandAugment (RandAug)\cite{randaug}, OnlineAugment (OnlineAug)\cite{onlineaugment}, and Adv. AA\cite{AdvAA}.

For CIFAR-10 and CIFAR-100, we used the smoothing parameter $\epsilon$ achieving the best error rate in the previous section.
We set $\epsilon$ to 0.1 for training on ImageNet, following the tendency in Fig. \ref{fig:ab_lbl_smoothing} (i.e., the smaller $\epsilon$ tends to be better for difficult tasks).

We show the comparison results in Tab. \ref{tab:comp}.
\begin{table*}[t]
\centering
\scalebox{0.9}{
\begin{tabular}{l|ccc|ccc|c}
                          & \multicolumn{3}{c|}{CIFAR-10} & \multicolumn{3}{c|}{CIFAR-100}&ImageNet\\
                          & WRN-28-10  & Shake-Shake & PyramidNet & WRN-28-10    & Shake-Shake & PyramidNet  & ResNet-50\\ \toprule[1.1pt]
Baseline                  &   3.9       &     2.9     &    2.7     &   18.8        &   17.1      &     14.0  & 23.7/6.9\\
Cutout\cite{cutout}      &   3.1       &     2.6     &    2.3     &   18.4        &   16.0      &     12.2  & - \\ 
AA\cite{AA}              &   2.6       &     2.0     &    1.5     &   17.1        &   14.3      &     10.7  & 22.4/6.4\\
PBA\cite{pba}            &   2.6       &     2.0     &    1.5     &   16.7        &   15.3      &     10.9  & -\\
RandAug\cite{randaug}    &   2.7       &     2.0     &    1.5     &   16.7        &   -      &    -   & 22.4/6.2\\
Fast AA\cite{FastAA}     &   2.7       &     2.0     &    1.7     &   17.3        &   14.6      &     11.7  & 22.4/6.3\\
Faster AA\cite{FasterAA} &   2.6       &     2.0     &      -     &   17.3        &   15.0      &     -     & 23.5/6.8\\
DADA\cite{dada}          &   2.7       &     2.0     &    1.7     &   17.5        &   15.3      &     11.2  & 22.5/6.5\\ 
OnlineAug\cite{onlineaugment} & 2.4 & - & - & 16.6 & - & - & 22.4/-\\
Adv. AA$^\dagger$~\cite{AdvAA}     &   (1.9)       &     (1.9)    &    (1.4)     &   (15.5)        &   (14.1)      &     (10.4) & (20.6/5.5)\\
Ours                      &   2.5       &     2.0     &    1.5     &   16.8        &   14.5      &     11.8   & 22.2/6.3\\ 
\end{tabular}
}
\caption{Test set top-1 error rates (\%) on CIFAR-10 and CIFAR-100 and validation set top-1/top-5 error rates on ImageNet. Please note that Adv. AA$^\dagger$ uses multiple augmented samples per mini-batch.}
\label{tab:comp}
\end{table*}
Our method achieves error rates comparable to other method except for Adv. AA, which uses multiple augmented samples per mini-batch.
In particular, our method achieves the lowest top-1 error rate on ImageNet among the methods without a multiple augmentation strategy.

\subsection{Semantic segmentation}
We also evaluate our method on Cityscapes\cite{cityscapes}.
We implement widely used models with a ResNet-101 backbone, FCN-32s\cite{fcn}, PSPNet\cite{pspnet}, and Deeplabv3\cite{deeplabv3}.
As baseline methods, we adopted RandAugment\cite{randaug} and TrivialAugment\cite{trivialaug}.
These methods also does not need careful parameter tuning because they have few or no hyperparameters.

We show the results in Tab. \ref{tab:semseg}.
\begin{table}[]
\centering
\scalebox{0.9}{
\begin{tabular}{c|cccc}
Model & Baseline  & RA\cite{randaug}  & TA\cite{trivialaug} & Ours \\ \toprule[1.1pt]
FCN-32 & 71.7  & 71.0 & 71.3 & \textbf{72.1} \\
PSPNet & 77.7 & \textbf{78.8} &  77.5 & \textbf{78.8} \\
Deeplabv3 & 78.4 & 79.3 & 78.9 & \textbf{79.4}     
\end{tabular}
}
\caption{Validation set mIoU (\%) on Cityscapes. RA and TA denote RandAugment\cite{randaug} and TrivialAugment\cite{trivialaug}, respectively.}
\label{tab:semseg}
\end{table}
Our method achieves the best mIoU for each model.
RandAugment hurts the mIoU for FCN-32s, and TrivialAugment does not improve the mIoU for any of the models.
In fact, it has been reported that TrivialAugment does not work well for tasks other than image classification\cite{trivialaug}.
We suspect that the search spaces of RandAugment and TrivialAugment are not suitable for semantic segmentation tasks and model capacities.
However, our method improves mIoU for all conditions without adjusting parameters from the classification tasks.

\subsection{Unsupervised representation learning}
Finally, we evaluate our method with unsupervised representation learning tasks using SimSiam\cite{simsiam}.
To generate two different views of the same image, we used two augmentation models, $a_{\phi_1}$ and $a_{\phi_2}$.
The detailed settings can be found in the appendix.

The results are shown in Tab. \ref{tab:uns}.
\begin{table}[]
    \centering
    \scalebox{0.9}{
    \begin{tabular}{c|ccc}
        \#Epochs               & 100  & 200  & 400\\ \toprule[1.1pt]
        Baseline~\cite{simsiam} & 68.1 & 70.0 & 70.8\\
        RandAugment~\cite{randaug}     & 68.0 & 70.0 & 70.7\\
        TrivialAugment~\cite{trivialaug}  & 62.7 & 68.7 & \textbf{71.3}\\
        Ours            & \textbf{68.2} & \textbf{70.2} & 71.0
    \end{tabular}
    }
    \caption{ImageNet linear evaluation accuracy (\%) for various pretraining epochs. We set batch size to 256 for pretraining.}
    \label{tab:uns}
\end{table}
TrivialAugment causes underfitting for 100 and 200 epoch training due to the diversity of augmentation induced by the strong randomness.
RandAugment does not contribute to the improvement of accuracy.
Our proposed method consistently improves the accuracy for all training schedules, although the improvements are slight.
This would be because the online optimization frameworks have aspects similar to curriculum learning~\cite{bengio2009curriculum}.
In other words, our method adjusts the augmentation magnitude according to the learning progress of the target model.

\section{Conclusion}
We proposed an online data augmentation optimization method called TeachAugment that introduces a teacher model into the adversarial data augmentation and makes it more informative without the need for careful parameter tuning.
We also proposed neural network based augmentation that simplified the search space design and enabled updating of the data augmentation with the gradient method.
In experiments, our method outperformed existing data augmentation search frameworks, including state-of-the-art methods, on image classification, semantic segmentation, and unsupervised representation learning tasks without adjusting the hyperparameters for each task.

\textbf{Limitation.}
The proposed color augmentation cannot represent transformation using global information of a target image, such as Equalize and AutoContrast, because of the lack of global information in the input.
Such transformations may be realized using the color histogram as the context vector, although at the expense of computational cost, especially for high-resolution images.
Moreover, we only focused on geometric and color augmentation, but there are many advanced augmentation that are not categorized with them, for example, cutout\cite{cutout} and mixup\cite{mixup}.
Considering such augmentations will be future work.

The training time for TeachAugment is approximately three times longer than the conventional training procedure because updating the augmentation model requires the forward and backward computation of the target model and the teacher model, in addition to updating the target network.
However, the training time of TeachAugment is almost the same as other online methods\cite{PA,onlineaugment} and is a realistic time in comparison to AutoAugment~\cite{AA}.
We believe that it is a negligible problem under the recent advances of computational power.

\begin{appendix}
\section{Training setups}
\subsection{Image classification}
We summarize the hyperparamters for training in Tab. \ref{tab:hyp}.
\begin{table*}[]
\begin{tabular}{c|ccccc}
                    & WideResNet-40-2 & WideResNet-28-10 & Shake-Shake (26 2$\times$96d) & PyramidNet & ResNet-50 \\ \toprule[1.1pt]
Learning rate       & 0.1             & 0.1             & 0.01                       & 0.05       & 0.05      \\
Weight decay        & 2e-4            & 5e-4            & 1e-3                       & 5e-5       & 1e-4      \\
Epochs              & 200             & 200             & 1,800                       & 1,800       & 270       \\
Batch size          & 128             & 128             & 128                        & 64         & 128       \\
Learning rate decay & cosine          & cosine          & cosine                     & cosine     & step     
\end{tabular}
\centering
\caption{Hyperparameters for classification tasks. We set parameters following \cite{FastAA}. Note that we show the batch size per GPU and the learning rate was multiplied by the number of GPUs (e.g., if two GPUs are used for training, the learning rate is doubled). We used a single GPU for WideResNet-40-2 and WideResNet-28-10, and four GPUs for the other models. We decayed the learning rate for ResNet-50 by 10-fold at epochs 90, 180, and 240.}
\label{tab:hyp}
\end{table*}
Each model was trained with Nesterov’s accelerated gradient method\cite{nesterov} in the stochastic gradient descent.
The cross entropy loss between the model prediction and the ground truth label was used as the loss function.
We gradually warmed up the learning rate for five epochs until it reached the predefined learning rate shown in Tab. \ref{tab:hyp}.
As baseline augmentation, we used random horizontal flipping and random cropping with a crop size of 32 for CIFAR-10 and CIFAR-100 and 224 for ImageNet.
In addition, we also used Cutout with a crop size of 16 for CIFAR-10 and CIFAR-100, and random color distortion for ImageNet.\footnote{The implementation of the baseline augmentation and models are based on \url{https://github.com/kakaobrain/fast-autoaugment}.}

\subsection{Semantic segmentation}
We trained all models using the stochastic gradient descent with momentum of 0.9 for 300 epochs.
The cross entropy loss between the model prediction and the ground truth label was used as the loss function.
We set the initial learning rate to 5e-3 and decayed it with poly learning rate decay where
the initial learning rate was multiplied by $\left(1-\frac{iter}{max\_iter}\right)^{0.9}$.
The coefficient of the auxiliary loss used in \cite{pspnet} was set to 0.4.
As baseline augmentation, we used random horizontal flipping and random cropping with a crop size of 1024.
For TeachAugment, the smoothing parameter $\epsilon$ in label smoothing was set to 0.1.
For RandAugment, the number of transformations $n$ and the magnitude $m$, were set to 1 and 5, which were tuned for FCN-32s using grid search.

\subsection{Unsupervised representation learning}
We evaluated each method following the linear evaluation setting\cite{simsiam}.\footnote{The experimental code is based on \url{https://github.com/facebookresearch/simsiam}.}
We modified the objective of our method for SimSiam as follows:
\begin{align}
\nonumber
\max_{\phi_1,\phi_2}\min_\theta\mathbb{E}_{x\sim\mathcal{X}}[&L(f_\theta(a_{\phi_1}(x)),f_\theta(a_{\phi_2}(x)))\\
&-L(f_\theta(a_{\phi_1}(x)),f_\theta(a_{\phi_2}(x)))],
\end{align}
where $L$ denotes the cosine distance.
Because the non-saturating loss and label smoothing cannot be applied to the cosine distance, we omitted them in this experiment.

Note that our method with the EMA teacher cannot be simply applied to other methods, such as BYOL\cite{byol} and MoCo\cite{moco,mocov2}, because they already integrate the EMA teacher into their training frameworks.
It will be future work to investigate combinations of such methods.

As pretraining, we trained ResNet-50 for 100, 200, and 400 epochs with a batch size of 256.
The momentum SGD was employed as the optimizer.
The learning rate and the momentum were set to 0.05 and 0.9, respectively.
After pretraining, we trained a linear classifier for 90 epochs with a batch size of 4,096.
We set the hyperparameters for each of the methods to the same as the parameters for ImageNet classification.
As baseline augmentation, we used the same augmentation as in SimSiam\cite{simsiam}, namely, random cropping, random horizontal flipping, color jittering, and Gaussian blur.
For more details and SimSiam's hyperparameters, please refer to \cite{simsiam}.

\section{Pseudo-Code of TeachAugment}
We show the pseudo-code of TeachAugment in Algorithm \ref{alg:ta}.
\begin{algorithm}[t]
\caption{Training procedure for TeachAugment}
\label{alg:ta}
\begin{algorithmic}[1]
\renewcommand{\algorithmicrequire}{\textbf{Input:}}
\renewcommand{\algorithmicensure}{\textbf{Output:}}
\REQUIRE A target model $f_\theta$, a teacher model $f_{\hat{\theta}}$, dataset $\mathcal{X}$, the number of inner iterations $n_\text{inner}$, learning rate $\eta_\theta$ and $\eta_\phi$, and decay rate for the EMA teacher $\xi$ 
\WHILE{$\theta$ has not converged}
\FOR{$i=0,\cdots,n_\text{inner}$}
\IF{$f_{\hat{\theta}}$ is the EMA teacher}
\STATE Update teacher weights, $\hat{\theta}\leftarrow \xi\hat{\theta}+(1-\xi)\theta$
\ENDIF
\STATE Randomly sample a mini-batch, $\{x^b\}^B_b\sim\mathcal{X}$
\STATE Compute loss for the target model,\\$L_\theta=\sum_bL(f_\theta(a_\phi(x^b)))$
\STATE Update $\theta$ by the gradient descent,\\
$\theta\leftarrow \theta-\eta_\theta\partial L_\theta/\partial\theta$
\ENDFOR
\STATE Randomly sample a mini-batch, $\{\bar{x}^b\}^B_b\sim\mathcal{X}$
\STATE Compute loss for the augmentation model,\\ $L_\phi=\sum_b(L(f_{\theta}(a_\phi(\bar{x}^b)))-L(f_{\hat{\theta}}(a_\phi(\bar{x}^b))))$
\STATE Update $\phi$ by the gradient ascent, $\phi\leftarrow \phi+\eta_\phi\partial L_\phi/\partial\phi$
\ENDWHILE
\end{algorithmic}
\end{algorithm}

\section{Details of augmentation model}
\label{sec:aug-model}
For the geometric augmentation, we used a three-layer perceptron.
The dimension of the noise vector was 128 and the number of units in hidden layers was 512.
As a non-linear activation function, we used leaky ReLU\cite{lrelu} with the negative slope of 0.2.
The output, $A^\mathrm{unnorm}$, was normalized through the sigmoid function:
\begin{align}
    A= \lambda_{g_\text{scale}}(\mathrm{sigmoid}(A^\mathrm{unnorm}) - 0.5),
\end{align}
where $\lambda_{g_\text{scale}}$ controls the search range of $A$, and we set it to 0.5 (i.e., $A\in(-0.25, 0.25)^{2\times3}$).

\begin{figure}
    \centering
    \includegraphics[clip,width=1\hsize]{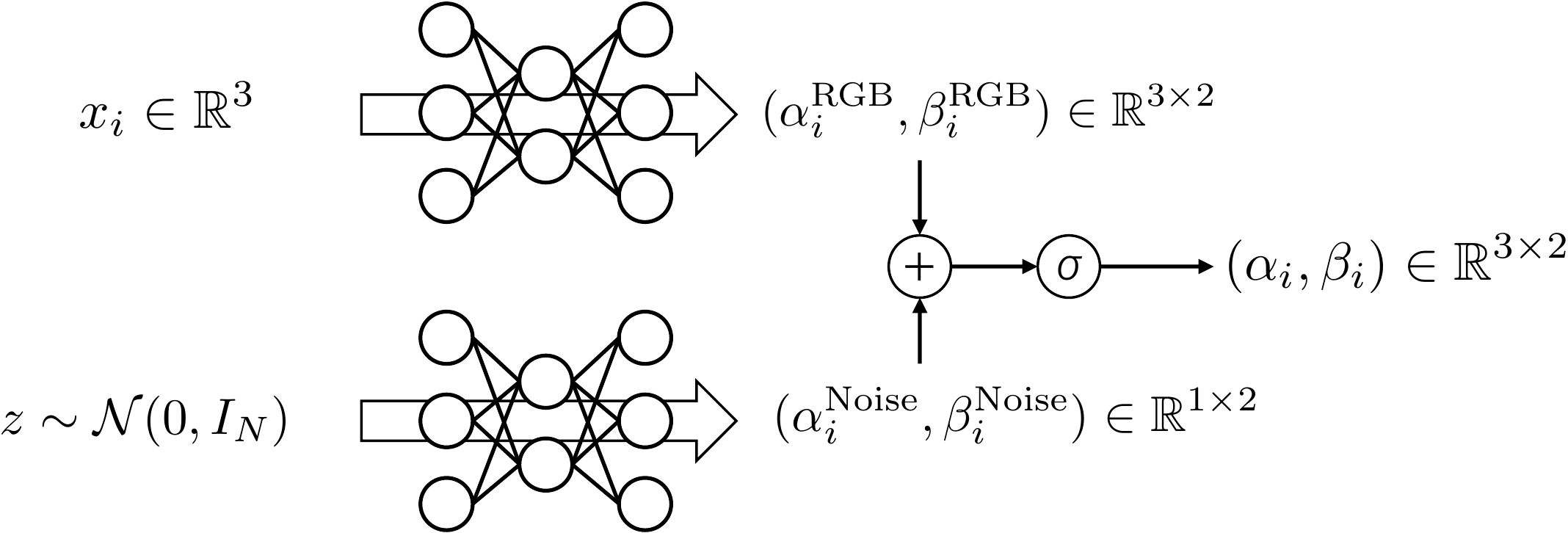}
    \caption{Illustration of the color augmentation model. $\sigma$ denotes the sigmoid function.}
    \label{fig:caug-comp}
\end{figure}

For the color augmentation, we used two three-layer perceptrons that receive an RGB vector and a noise vector as inputs and add up their outputs.
The number of units in hidden layers was 128 and 512, respectively.
As the non-linear activation, leaky ReLU was used.
We illustrate the computational scheme of the color augmentation model in Fig. \ref{fig:caug-comp}.
The former model outputs 3-dimensional scale and shift parameters:
\begin{align}
    (\alpha_i^\text{RGB},\beta_i^\text{RGB})\in\mathbb{R}^{3\times2},
\end{align}
and the latter outputs scalar scale and shift parameters:
\begin{align}
    (\alpha_i^\text{Noise},\beta_i^\text{Noise})\in\mathbb{R}^{1\times2}.    
\end{align}
The scale and shift parameters from the noise vector control the global brightness of images.
Then, we add up these scale and shift parameters:
\begin{align}
    (\alpha_i^\mathrm{unnorm})_j=(\alpha_i^\text{RGB})_j+\alpha_i^\text{Noise},\\
    (\beta_i^\mathrm{unnorm})_j=(\beta^\text{RGB}_i)_j+\beta^\text{Noise}_i,
\end{align}
where $(\alpha_i)_j$ denotes the $j$-th element of $\alpha_i\in\mathbb{R}^3$.
We normalized the scale and shift parameters, $(\alpha_i^\mathrm{unnorm},\beta_i^\mathrm{unnorm})$, using the sigmoid function:
\begin{align}
    \alpha_i=\lambda_{c_\text{scale}}(\mathrm{sigmoid}(\alpha_i^\mathrm{unnorm})-0.5) + 1,\\
    \beta_i=\lambda_{c_\text{scale}}(\mathrm{sigmoid}(\beta_i^\mathrm{unnorm})-0.5),
\end{align}
where $\lambda_{c_\text{scale}}$ controls the search range of $\alpha_i$ and $\beta_i$, and we set it to 0.8, namely, $\alpha_i\in(0.6,1.4)$ and $\beta_i\in(-0.4,0.4)$.

We adopted AdamW~\cite{adamw} as the optimizer for the augmentation model.
The learning rate and the weight decay were set to 1e-3 and 1e-2, which are the default parameters in PyTorch\cite{pytorch}.
Dropout\cite{dropout} was applied after the linear layers except for the output layer with the drop ratio of 0.8.

\section{Learning pipeline of probabilities $p_c$ and $p_g$}
We made the decision process of applying the augmentation differentiable using weights sampled from the relaxed Bernoulli distribution defined in \cite{concrete}.\footnote{The relaxed Bernoulli distribution is referred to as the BinConcrete distribution in \cite{concrete}.}
A sample from the relaxed Bernoulli, $w\sim\mathrm{ReBern}(p,\tau)$, is obtained as follows:
\begin{align}
    \label{eq:sample-rebern}
    w=\mathrm{sigmoid}((L+\log p)/\tau),\ L\sim\mathrm{Logistic}(0,1),
\end{align}
where $\tau$ and $p$ denote the temperature parameter and a probability that corresponds to $p_c$ and $p_g$ in our case and $L$ is a sample from the Logistic distribution, which is obtained by $L=\log(U)-\log(1-U),\ U\sim\mathrm{Uniform}(0,1)$.
We set $\tau$ to 0.05 following \cite{FasterAA}.
We note that the sampling procedure, Eq. \ref{eq:sample-rebern}, is differentiable with respect to the probability $p$.

We compute weighted sum of the parameters generated by augmentation models and parameters that make the augmentation the identity mapping:
\begin{align}
    \hat{\alpha}_i = w_c \alpha_i + (1-w_c)\cdot1,\\
    \hat{\beta}_i=w_c\beta_i+(1-w_c)\cdot0,\\
    \hat{A}=w_gA+(1-w_g)I,
\end{align}
where $w_c\sim\mathrm{ReBern}(p_c,\tau)$ and $w_g\sim\mathrm{ReBern}(p_g,\tau)$.
We use $(\hat{\alpha}_i,\hat{\beta}_i)$ and $\hat{A}$ to transform images, instead of $(\alpha_i,\beta_i)$ and $A$:
\begin{align}
    \tilde{x}_i=t(\hat{\alpha}_i\odot x_i+\hat{\beta}_i),\\
    \hat{x}=\mathrm{Affine}(\tilde{x},\hat{A}+I).
\end{align}
Because the sampling procedure from $\mathrm{ReBern}(p,\tau)$ is differentiable with respect to $p$, we can also update the probabilities using the gradient method.

\section{Additional results}
\subsection{Relation between consistency regularization and TeachAugment}
TeachAugment can be viewed as a method that minimizes the distance between predictions of the target model and the teacher model.
We show it qualitatively.

\begin{figure}
    \centering
    \includegraphics[clip,width=1\hsize]{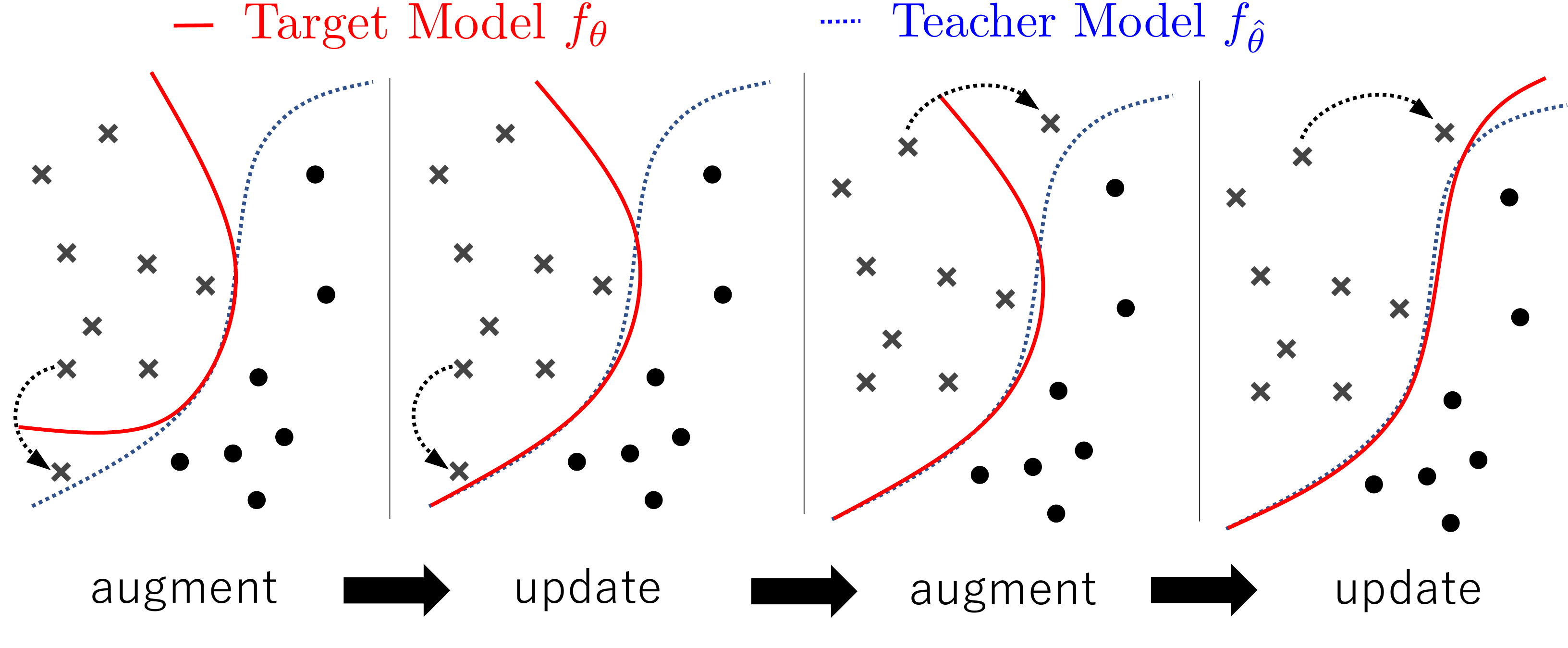}
    \caption{Illustration of the training process in TeachAugment. The data augmentation transforms a data point to the point so that $L(f_\theta(a_\phi(x)))>L(f_{\hat{\theta}}(a_\phi(x)))$. Then, the target model is updated to minimize loss for the augmented point. By repeating this process, the decision boundary of the target model is close to that of the teacher model.}
    \label{fig:updating-pro}
\end{figure}

We assume that data augmentation $a_\phi$ can transform an input data point to any point in the input space, and the teacher model is fixed during training.
Then, solving $\max_\phi L(f_\theta(a_\phi(x)))- L(f_{\hat{\theta}}(a_\phi(x)))$ corresponds to searching the data augmentation maximizing the distance between the predictions under the constraint of $L(f_\theta(a_\phi(x)))> L(f_{\hat{\theta}}(a_\phi(x)))$.
Then, the target model updates its decision boundary for minimizing the loss for the augmented data.
This procedure is illustrated in Fig. \ref{fig:updating-pro}.
If the decision boundary of the target model completely corresponds to that of the teacher model, the data augmentation reaches the stationary point because the objective $L(f_\theta(a_\phi(x)))- L(f_{\hat{\theta}}(a_\phi(x)))$ is 0 for all data points.

However, the target model would not match the teacher model in practice, because above analysis lacks certain points:
(1) The data augmentation cannot transform an input data point to any point in the input.
(2) We used the non-saturating loss for the image classification tasks instead of cross entropy. Thus, $L(f_\theta(a_\phi(x)))- L(f_{\hat{\theta}}(a_\phi(x)))\neq0$ if the decision boundary of the target model is the same as that of the teacher model.
(3) TeachAugment does not explicitly minimize the distance between the predictions.
In particular, (3) is a more critical point compared to (1) and (2) because they may be solved by adopting a model large enough to satisfy the requirement as the augmentation model, and using an original loss function instead of the non-saturating loss.
Because of the lack of explicit costs for the consistency, TeachAugment does not ensure that the target model converges with the teacher model.

However, we believe that the above analysis gives some insights, and comparing TeachAugment to consistency regularization is important.
Thus, we trained model to minimize the following costs and compared the results to TeachAugment:
\begin{align}
    \label{eq:consis}
    \min_\theta L(f_\theta(x))+D(f_\theta(x),f_{\hat{\theta}}(x)),
\end{align}
where $D(\cdot,\cdot)$ denotes the distance function.
We used the mean squared error and Kullback–Leibler divergence as $D(\cdot,\cdot)$.
These functions are widely used in the consistency regularization and the knowledge distillation.
We refer to the former as MSE Consistency and the later as KLD Consistency.
Note that we use random horizontal flipping, random cropping, and cutout\cite{cutout} as the data augmentation but omitted them in Eq. \eqref{eq:consis} because they were not optimized.
As the teacher model, we used the same EMA teacher in TeachAugment.

The results are shown in Tab. \ref{tab:consistency}.
\begin{table}[]
    \centering
    \begin{tabular}{c|cc}
        Dataset          & CIFAR-10 & CIFAR-100 \\ \toprule[1.1pt]
        Baseline         & 3.1      & 18.4 \\
        KLD consistency  & 3.1      & 18.2 \\
        MSE consistency  & 3.2      & 18.5 \\
        TeachAugment     & \textbf{2.5} & \textbf{16.8} \\
    \end{tabular}
    \caption{Comparison with the consistency regularization. We report the error rates of WideResNet-28-10. For training with KLD and MSE consistency, we used random horizontal flipping, random cropping, and cutout\cite{cutout} as data augmentation.}
    \label{tab:consistency}
\end{table}
The consistency methods do not improve the error rates from the baseline.
Thus, TeachAugment has different properties than the consistency regularization, and it works well in supervised learning.

\subsection{Qualitative analysis}
We show augmented images obtained by the augmentation model trained with various objective functions: TeachAugment, Adversarial AutoAugment (Adv. AA)\cite{AdvAA}, and PointAugment\cite{PA}.
All methods used the same proposed augmentation model as data augmentation.

The objective of Adv. AA is as follows:
\begin{align}
    \max_\phi\min_\theta L(f_\theta(a_\phi(x))).
\end{align}
Also, the objective of PointAugment is as follows:
\begin{align}
    &\min_\theta L(f_\theta(a_\phi(x))),\\
    \nonumber
    \min_\phi L(f_\theta(a_\phi(x)))&\\
    +|1-\exp(&L(f_\theta(a_\phi(x)))-\rho L(f_\theta(x)))|,
\end{align}
where $\rho$ is a dynamic parameter defined as $\rho=\exp(y^T\cdot f_\theta(a_\phi(x)))$.
Note that Adv. AA updates augmentation functions using the REINFORCE algorithm\cite{reinforce} because many functions in the search space of AutoAugment are non-differentiable.
However, in our experiments, because our proposed augmentation is differentiable, we updated augmentation with the gradient descent rather than the REINFORCE algorithm.

\begin{figure*}
    \centering
    \begin{tabular}{cccc}
    \includegraphics[width=0.2\hsize]{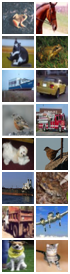} &
    \includegraphics[width=0.2\hsize]{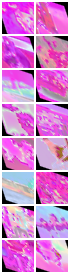} &
    \includegraphics[width=0.2\hsize]{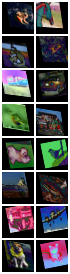} &
    \includegraphics[width=0.2\hsize]{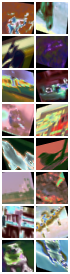} \\
    (a) Original Image & (b) Adv. AA & (c) PointAugment & (d) Ours
    \end{tabular}
    \caption{Augmented images obtained with CIFAR-10 using various methods..}
    \label{fig:c10-aug}
\end{figure*}

\begin{figure*}
    \centering
    \begin{tabular}{cccc}
    \includegraphics[width=0.2\hsize]{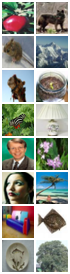} &
    \includegraphics[width=0.2\hsize]{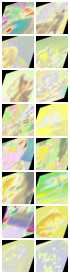} &
    \includegraphics[width=0.2\hsize]{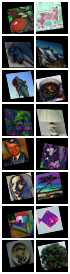} &
    \includegraphics[width=0.2\hsize]{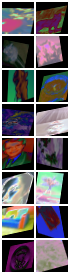} \\
    (a) Original Image & (b) Adv. AA & (c) PointAugment & (d) Ours
    \end{tabular}
    \caption{Augmented images obtained with CIFAR-100 using various methods..}
    \label{fig:c100-aug}
\end{figure*}

The augmented images of CIFAR-10 and CIFAR-100 are shown in Figs. \ref{fig:c10-aug} and \ref{fig:c100-aug}.
The images augmented by Adv. AA obviously collapse, so the meaningful information is lost.
The augmented images obtained by PointAugment and the proposed method are recognizable, but the proposed method transforms images more strongly than PointAugment so that the proposed method distorts the aspect ratio for the geometric augmentation.
PointAugment binds the difficulty of the augmented images through a dynamic parameter $\rho \leq \exp(1)$, but the proposed method requires only that the augmented images are recognizable for the teacher model.
As a result, the proposed method allows stronger augmentation than PointAugment.

\subsection{Example of augmented images}
\begin{figure*}
    \centering
    \begin{tabular}{cc}
        \includegraphics[clip,width=0.4\hsize]{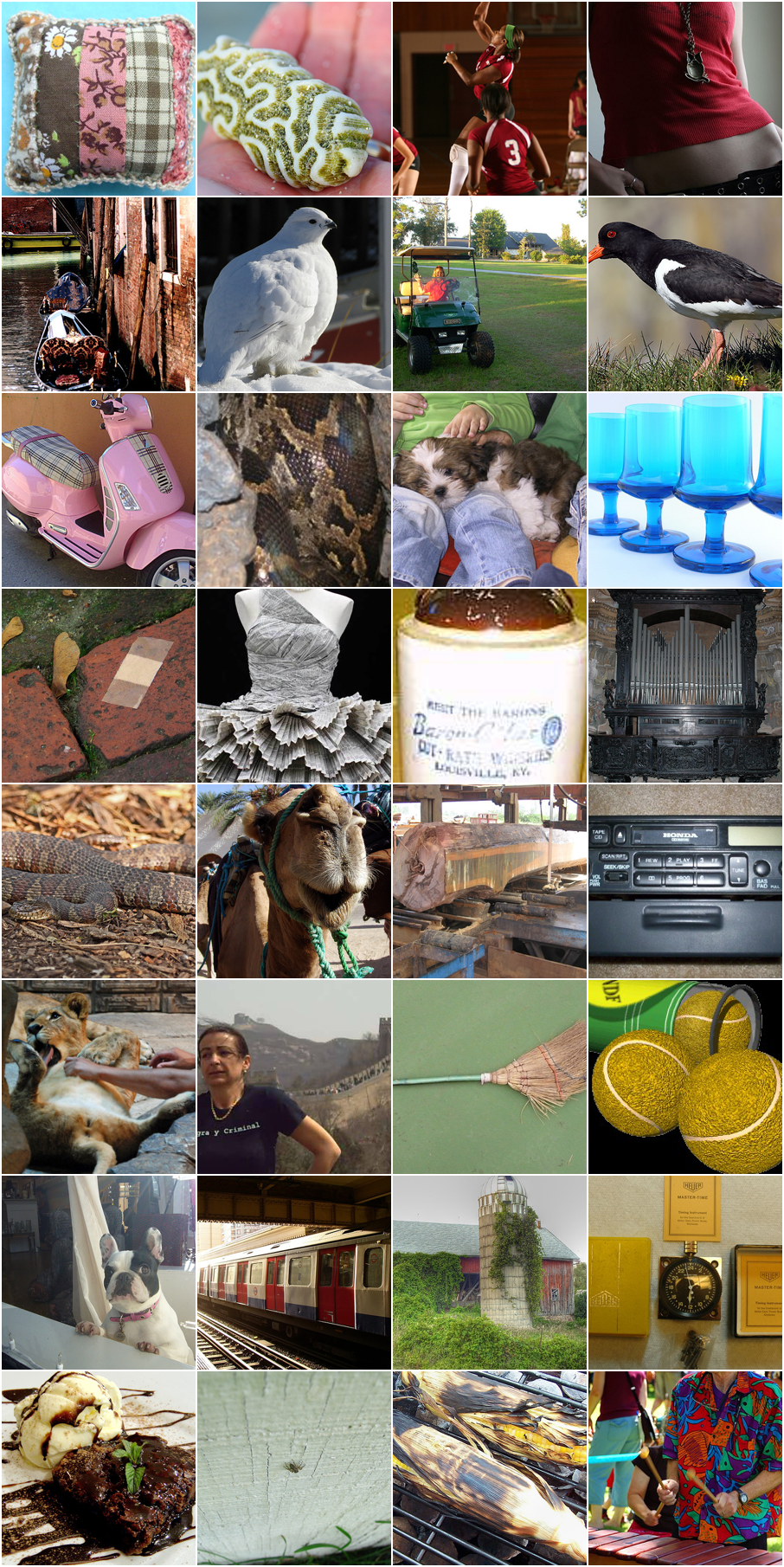} & 
        \includegraphics[clip,width=0.4\hsize]{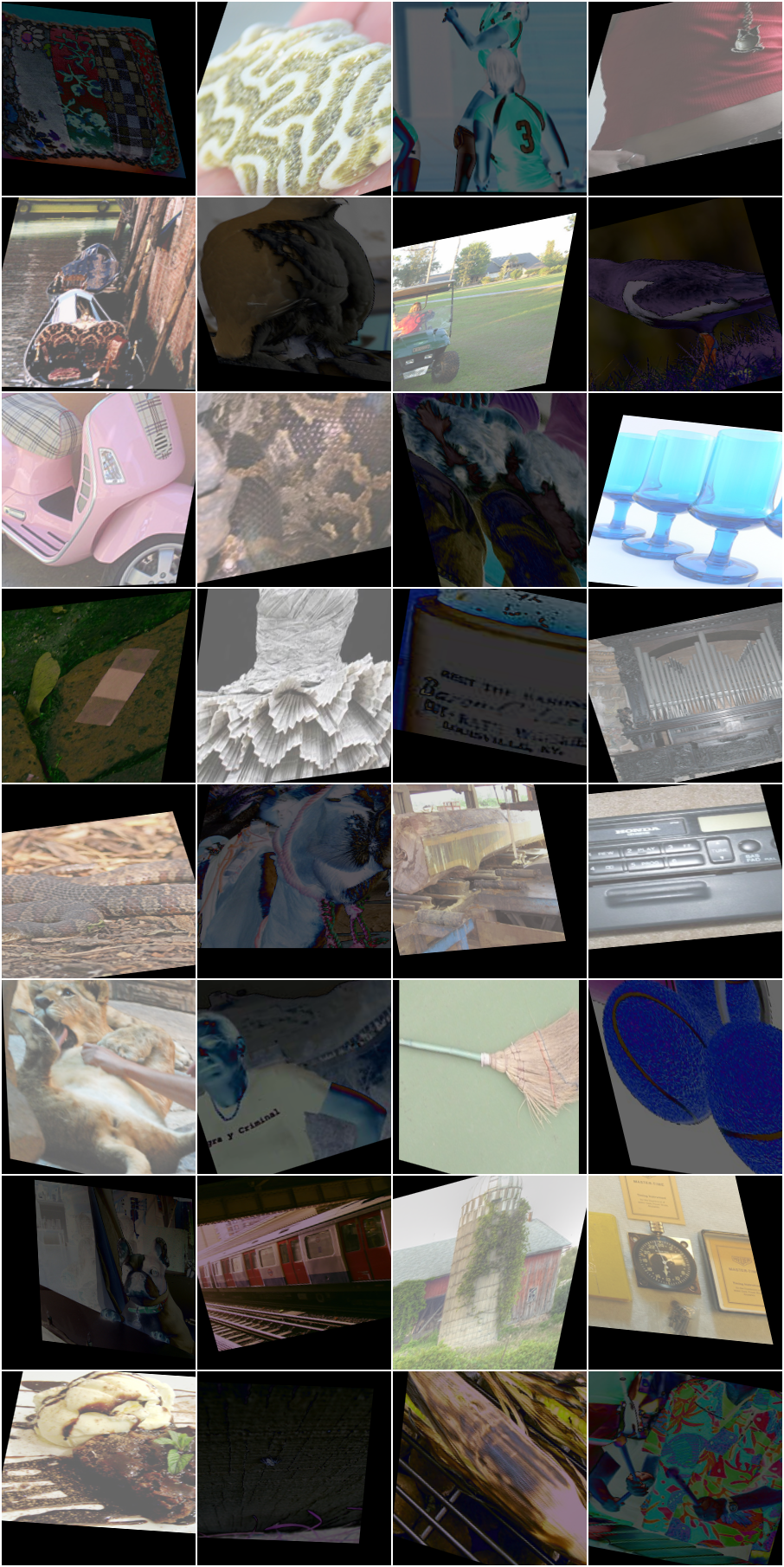} \\
        (a) Original images & (b) Augmented images
    \end{tabular}
    \caption{Augmentations obtained with ImageNet.}
    \label{fig:imagenet-aug}
\end{figure*}

\begin{figure*}
    \centering
    \begin{tabular}{cccc}
        \includegraphics[clip,width=0.2\hsize]{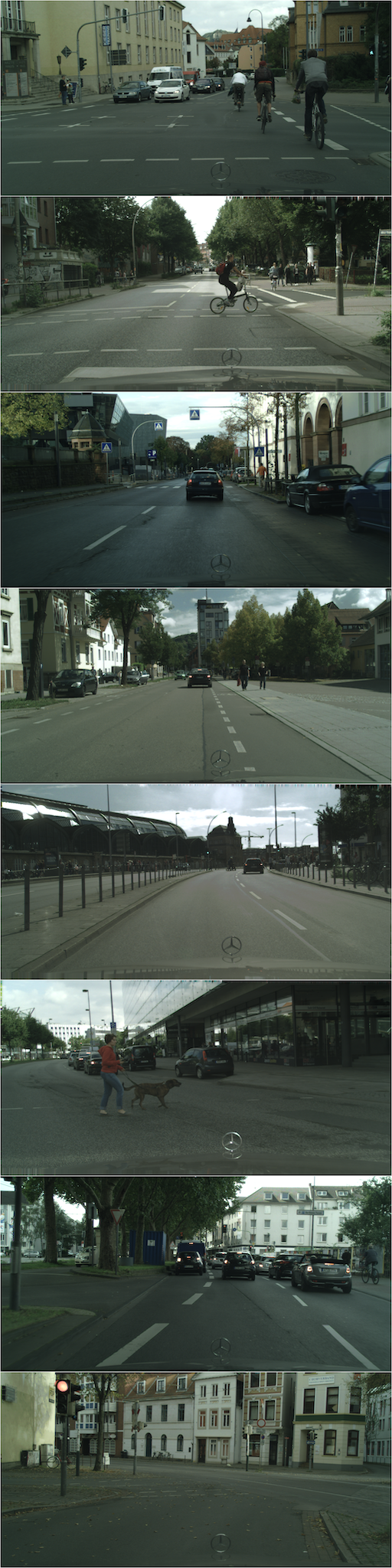} & 
        \includegraphics[clip,width=0.2\hsize]{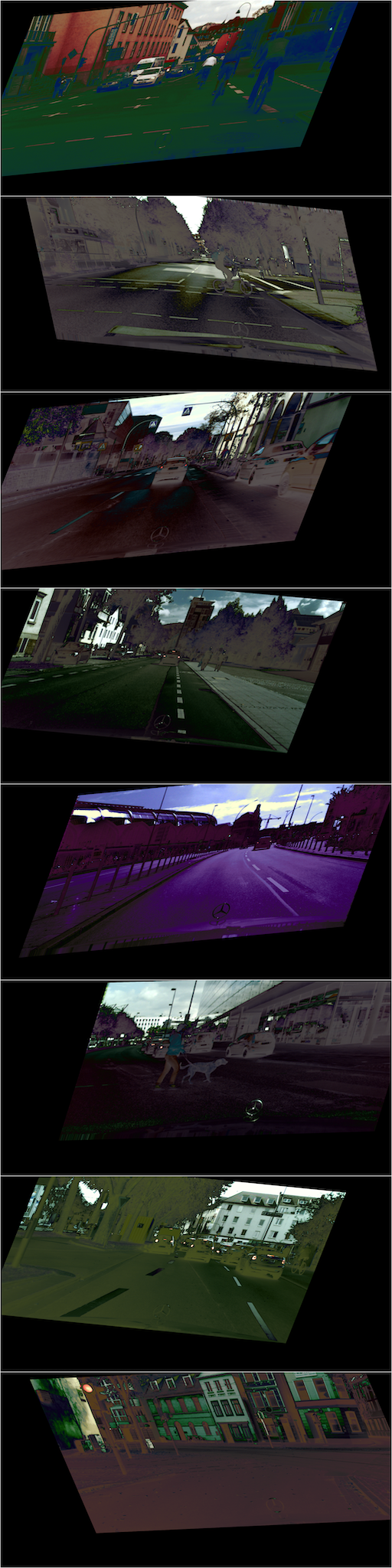} &
        \includegraphics[clip,width=0.2\hsize]{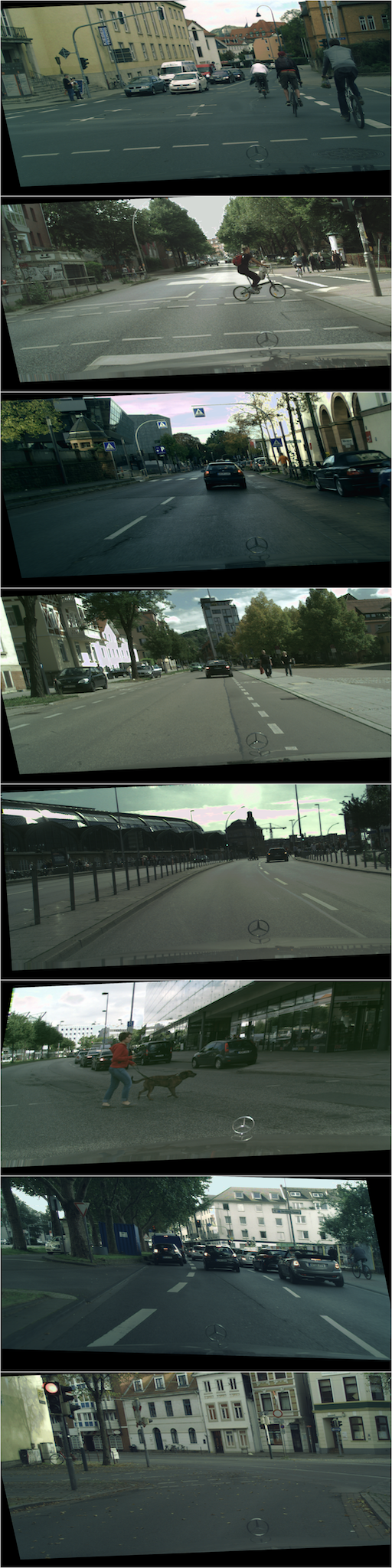} &
        \includegraphics[clip,width=0.2\hsize]{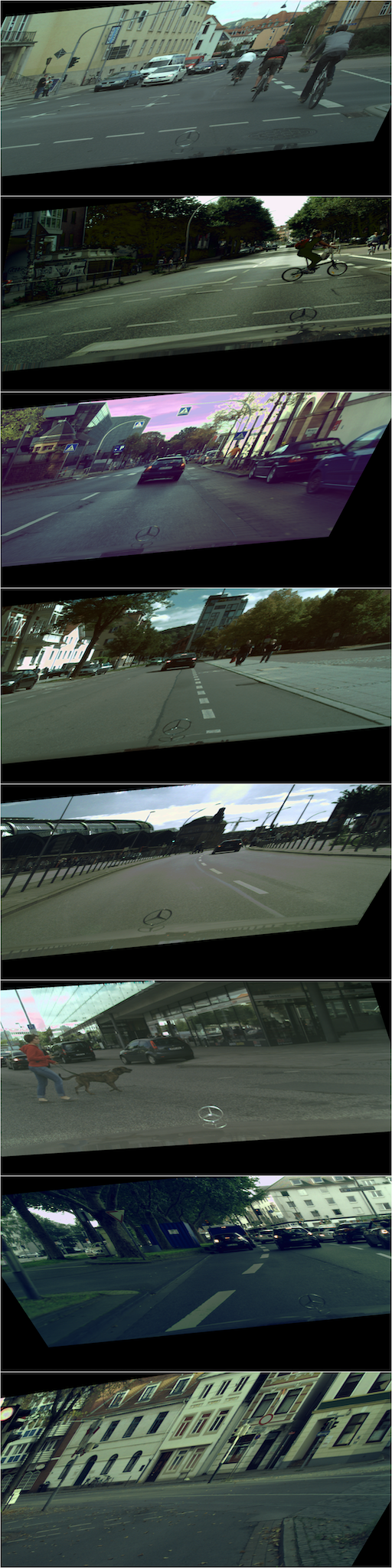} \\
        (a) Original images & (b) FCN-32s & (c) PSPNet & (d) Deeplabv3
    \end{tabular}
    \caption{Augmentations obtained with Cityscapes.}
    \label{fig:cityscapes-aug}
\end{figure*}

We show example results of augmentation for ImageNet and Cityscapes in Figs. \ref{fig:imagenet-aug} and \ref{fig:cityscapes-aug}.
As can see from Fig. \ref{fig:cityscapes-aug}, the augmentation obtained for FCN-32s is obviously different from the others.
Because the output stride of FCN-32s (i.e., the ratio of input image spatial resolution to final output resolution) is lower than that of the others, FCN-32s will have different properties than PSPNet and Deeplav3.
We believe that the difference leads the different augmentation of these models.
Moreover, it also leads the degradation of mIoU for FCN-32s in RandAugment and TrivialAugment.

\end{appendix}

{\small
\bibliographystyle{ieee_fullname}
\bibliography{egbib}
}

\end{document}